\documentclass[11pt,a4paper]{article}
\usepackage[hyperref]{rocling2025}
\usepackage{times}
\usepackage{latexsym}

\usepackage{booktabs}
\usepackage{float}
\usepackage{placeins}
\usepackage{amssymb}
\usepackage{amsmath}
\usepackage{pifont} 
\usepackage{dblfloatfix}
\usepackage{svg}
\usepackage{multirow}
\newcommand{\cmark}{\textcolor{green!60!black}{\ding{51}}}
\newcommand{\xmark}{\textcolor{red}{\ding{55}}} 
\definecolor{darkgreen}{RGB}{0,150,0}
\definecolor{darkred}{RGB}{200,0,0} 

\usepackage{microtype}

\roclingfinalcopy 

\title{A Preliminary Study of RAG for Taiwanese Historical Archives}

\author{
  Claire Lin\textsuperscript{1*}, Bo-Han Feng\textsuperscript{2*}, 
  Xuanjun Chen\textsuperscript{3*}, Te-Lun Yang\textsuperscript{4} \\
  \textbf{Hung-yi Lee\textsuperscript{3}, Jyh-Shing Roger Jang\textsuperscript{2,4}} \\
  \textsuperscript{1}Department of Information Management, National Taiwan University \\
  \textsuperscript{2}Department of Computer Science and Information Engineering, National Taiwan University \\
  \textsuperscript{3}Graduate Institute of Communication Engineering, National Taiwan University \\
  \textsuperscript{4}Graduate Institute of Networking and Multimedia, National Taiwan University \\
  {\small\texttt{\{b10705004, b10902031, d12942018, d12944007\}@ntu.edu.tw}} \\
  {\small\texttt{hungyilee@ntu.edu.tw, jang@mirlab.org}}
}

\begin{document}
\maketitle

\def\thefootnote{*}\footnotetext{Equal contribution.}\def\thefootnote{\arabic{footnote}}

\begin{abstract}
Retrieval-Augmented Generation (RAG) has emerged as a promising approach for knowledge-intensive tasks. However, few studies have examined RAG for Taiwanese Historical Archives. 
In this paper, we present an initial study of a RAG pipeline applied to two historical Traditional Chinese datasets, Fort Zeelandia and the Taiwan Provincial Council Gazette, along with their corresponding open-ended query sets. 
We systematically investigate the effects of query characteristics and metadata integration strategies on retrieval quality, answer generation, and the performance of the overall system. 
The results show that early-stage metadata integration enhances both retrieval and answer accuracy while also revealing persistent challenges for RAG systems, including hallucinations during generation and difficulties in handling temporal or multi-hop historical queries.
\end{abstract}


\section{Introduction}
Recent advances in large language models have substantially improved open-domain question answering and knowledge-intensive tasks. Retrieval-Augmented Generation (RAG)~\cite{lewis2020retrieval}, which combines document retrieval with text generation, has shown promise in mitigating hallucination and improving factuality. Prior research has primarily focused on English \citep{bajaj2018msmarcohumangenerated, kwiatkowski-etal-2019-natural, yang2024cragcomprehensiverag} or Simplified Chinese datasets \citep{lyu2024crudragcomprehensivechinesebenchmark, li2024cmmlumeasuringmassivemultitask} and general-purpose domains such as Wikipedia or web-collected questions.

However, much less attention has been given to RAG performance on underrepresented languages and culturally specific corpora, particularly in the humanities. Historical contexts in Traditional Chinese pose unique challenges, including unstructured documents, time-sensitive content, and linguistic differences between queries and archival sources. These factors complicate both retrieval and generation, making it unclear how well current RAG systems handle such materials.

To address this gap, we propose two Taiwanese historical datasets, Fort Zeelandia and Taiwan Provincial Council Gazette (TPCG), along with their associated query sets, as case studies for historical open-ended question answering. 
The datasets are annotated with query-level and document-level metadata, enabling fine-grained experiments on how query types and metadata integration strategies affect RAG performance. 
Through systematic evaluation across multiple retrieval methods and query characteristics, we demonstrate that early-stage metadata integration substantially improves system effectiveness. 
Furthermore, our findings reveal persistent challenges: hallucinations remain a recurring issue during generation, and questions involving temporal reasoning exhibit notable difficulty. Furthermore, our analysis of retrieval performance reveals that multi-hop and time-sensitive queries tend to yield lower recall, whereas early-stage metadata integration consistently delivers the strongest overall retrieval effectiveness.
\section{Related Work}
\begin{table*}[ht]
\centering
\small
\begin{tabular*}{\linewidth}{@{\extracolsep{\fill}}lccccc}
\toprule
\textbf{Dataset} & \textbf{Language} & \textbf{Humanities} & \textbf{Query-Passage Pairs} & \textbf{Metadata} \\
\midrule
MS MARCO         & English & \xmark & \cmark & Limited \\
Natural Questions& English & \xmark & \cmark & \xmark \\
MMLU             & English & \cmark & \xmark & \xmark \\
CMMLU             & Simplified Chinese & \cmark & \xmark & \xmark \\
Fort Zeelandia Query Set (Our) & Traditional Chinese & \cmark & \cmark & \cmark \\
TPCG Query Set (Our)           & Traditional Chinese & \cmark & \cmark & \cmark \\
\bottomrule
\end{tabular*}
\caption{Comparison of datasets by language, domain knowledge, structure, and metadata. Fort Zeelandia and TPCG Query sets stand out for their rich metadata and grounding in historical or contextual knowledge.}
\label{tab:dataset_comparison}
\end{table*}

RAG \citet{lewis2020retrieval} improves language model performance on knowledge-intensive tasks by incorporating relevant external information during generation. By grounding outputs in retrieved evidence, RAG reduces hallucinations when models encounter unfamiliar topics and alleviates the substantial cost of continuously retraining models to incorporate new knowledge.

Early benchmarks of RAG mainly relied on general-purpose datasets such as MS MARCO \citep{bajaj2018msmarcohumangenerated} and Natural Questions \citep{kwiatkowski-etal-2019-natural}. More recently, researchers have introduced domain-specific datasets in areas including biomedicine \citep{xiong-etal-2024-benchmarking,li2024biomedragretrievalaugmentedlarge,he2025retrievalaugmentedgenerationbiomedicinesurvey}, law \citep{pipitone2024legalbenchragbenchmarkretrievalaugmentedgeneration, Zheng_2025, 10887211}, and non-English languages such as Traditional Chinese \citep{yang2025knowledgeretrievalbasedgenerative}. However, RAG applications in the humanities are underexplored, particularly for Taiwanese historical materials.

Table \ref{tab:dataset_comparison} compares the key differences of existing benchmarks with the query sets from our newly introduced Fort Zeelandia and TPCG datasets. 
Firstly, in terms of humanities coverage, MS MARCO and Natural Questions primarily target general-purpose or factual QA and contain little to no humanities material, whereas MMLU \cite{hendrycks2021measuringmassivemultitasklanguage} and CMMLU \cite{li2024cmmlumeasuringmassivemultitask} include partial coverage through their broader topical scope. By contrast, our Fort Zeelandia and TPCG query sets are explicitly designed around humanities data, with a particular emphasis on historical materials. 
Secondly, with respect to query–passage alignment, MS MARCO and Natural Questions are constructed around paired queries and passages, a design we also adopt for Fort Zeelandia and TPCG query sets to support retrieval-based evaluation. MMLU and CMMLU, in contrast, rely on multiple-choice formats. 
Finally, in terms of metadata, our proposed datasets provide rich query- and document-level annotations, enabling more fine-grained retrieval experiments and analysis than existing resources.

\section{Dataset}
We introduce two Traditional Chinese datasets from Taiwanese historical archives: Fort Zeelandia and Taiwan Provincial Council Gazette (TPCG). 
We refer to the associated queries as the Fort Zeelandia Query Set and the TPCG Query Set, and to Fort Zeelandia and TPCG themselves as the document datasets in this paper. 

\subsection{Fort Zeelandia}
\begin{table}[ht]
\small
\centering
\begin{tabular*}{\columnwidth}{@{\extracolsep{\fill}}lccc}
    \toprule
    \textbf{Entity} & \textbf{Single-hop} & \textbf{Multi-hop} & \textbf{Total} \\
    \midrule
    Event        & 32 & 18 & 50 \\
    Item         & 14 &  2 & 16 \\
    People       & 19 &  4 & 23 \\
    Place        & 16 &  6 & 22 \\
    Time         & 19 &  4 & 23 \\
    Multi-entity & 0 &  39 & 39 \\
    \midrule
    \textbf{Total} & 100 & 73 & 173 \\
    \bottomrule
\end{tabular*}
\caption{Fort Zeelandia Query Set Entity Focus Distribution across Question Complexity}
\end{table}

This dataset is constructed from historical diaries\footnote{\href{https://taco.ith.sinica.edu.tw/tdk/}{https://taco.ith.sinica.edu.tw/tdk/}} documenting Dutch colonization of Taiwan in the 17th century. We collected 5,443 passages and collaborated with students from the Department of History, who created 173 queries and annotated the relevant passages for each query.

\textbf{Query-level Metadata.}
Each QA pair is annotated with query-level metadata, including:
\begin{itemize}
    \item \textbf{Question complexity}: Single-hop or multi-hop question. A multi-hop question requires combining information from multiple passagesto determine the answer, whereas a single-hop question can be answered using just one passage.
    \item \textbf{Entity focus}: Whether the question centers on a person, item, time, event, or location.
\end{itemize}
An example from the Fort Zeelandia dataset is demonstrated in Appendix \ref{fort_data}.

\subsection{Taiwan Provincial Council Gazette}
 The TPCG dataset comprises official meeting records from the Taiwan Provincial Council Assembly \footnote{\href{https://drtpa.th.gov.tw/index.php?act=Archive}{https://drtpa.th.gov.tw/index.php?act=Archive}}, spanning the mid to late 20th century, totaling 228,135 documents. To build the question answering benchmark, history students manually crafted 56 question-passage pairs based on selected gazette excerpts. The resulting dataset captures realistic information needs and research scenarios commonly encountered in historical inquiry.
\\
\textbf{Document-level Metadata.} \
TPCG is characterized with well-defined document-level metadata, enabling experiments on how structured context can be used to improve system performance. Each document is associated with:
\begin{itemize}
    \item \textbf{Time/Event Information}: Includes time information such as the start and end dates, volume and published date.
    \item \textbf{Person/Organization Information}: Covers participating members, agencies, decree, presiding officials and president at that time.
    \item \textbf{Content/Document Information}: Includes document title, abstract, content type, category, subject, keywords, attachments, references, and remarks.
\end{itemize}
An example from the TPCG dataset is demonstrated in Appendix \ref{tpcg_data}.

\section{Methods}

\begin{figure*}[t]
    \centering
\includegraphics[width=\linewidth]{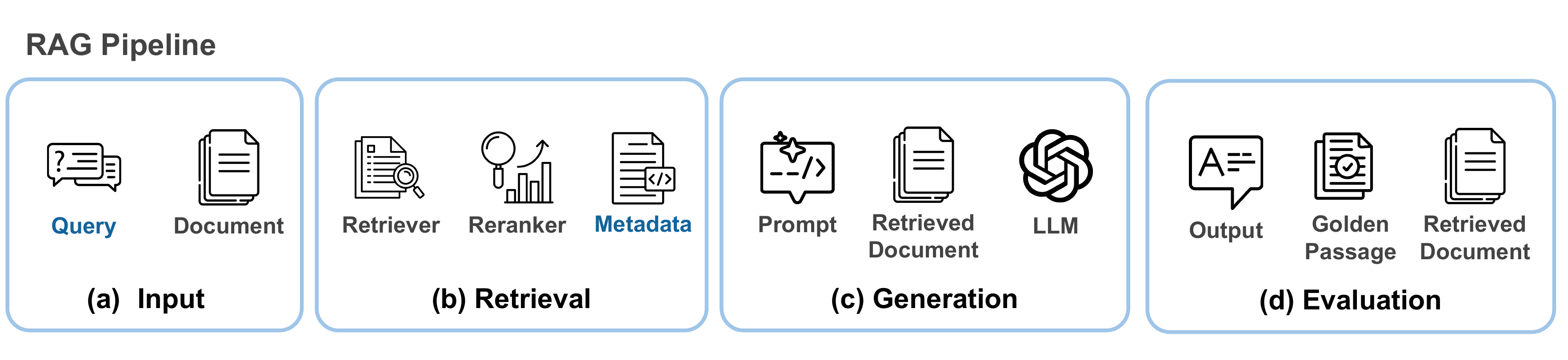}
    \caption{Overview of RAG pipeline and components in each stage. The two highlighted elements: \textbf{Query} and \textbf{Metadata} are the key factors that impact RAG system performance we focused on in this paper. The details of these factors are discussed in Section 3.1 and Section 3.2, respectively. Section 6.2 and Section 6.3 elaborates how these factors impact retrieval and generation performance.}
    \label{fig:rag_overview}
\end{figure*}

The RAG pipeline in Figure \ref{fig:rag_overview} comprises four stages: Input, Retrieval, Generation, and Evaluation. Throughout the pipeline, we (a) construct datasets and annotate query–passage pairs, (b) retrieve candidate passages using lexical, dense, and hybrid methods with optional metadata integration and reranking, (c) prompt a generator LLM with the query, retrieved passages, and metadata to generate an answer, and (d) assess answer quality with an LLM-as-judge protocol.


\subsection{Input}

The input stage in Figure \ref{fig:rag_overview} (a) covers data acquisition and annotation. We first crawl and normalize raw materials into document collections for Fort Zeelandia and TPCG datasets. Domain experts (Taiwanese history students) then author queries and annotate the associated gold passages, yielding high-quality query–passage pairs for RAG experimentation. To enable controlled analysis, we further annotate (i) question complexity (single-hop vs. multi-hop) and entity focus (people, event, time, place, item, or multi-entity) for Fort Zeelandia, and (ii) document-level metadata for TPCG, grouped into Time/Event, Person/Organization, and Document/Content categories. 


\subsection{Retrieval}

Given a user query, the retrieval stage in Figure \ref{fig:rag_overview} (b) identifies a small set of passages most likely to support grounded answer generation. This stage is essential in a RAG pipeline because it (i) grounds the generator in verifiable evidence to reduce hallucinations, (ii) filters a large corpus into a compact candidate set that fits the context window, and (iii) adapts to lexical, semantic information, and structured metadata in Fort Zeelandia and TPCG. The stage comprises two parts: retrieval models (sparse, dense, hybrid) that score query–passage relevance, and retrieval strategies that optionally use document-level metadata and a second-stage reranker. Together, these components return top-$k$ passages for the generation stage.


\subsubsection{Retrieval Models}

We instantiate three families of retrieval models:

\textbf{Sparse retrieval}. We adopt BM25 \citep{article}, which retrieves documents based on term-matching style term-frequency and inverse document frequency (TF-IDF) weighting \citep{10.5555/866292}, together with sparse embeddings derived from a BGE-M3-based model \citep{chen2024bgem3embeddingmultilingualmultifunctionality}.

\textbf{Dense retrieval}. A BGE-M3–based dense encoder maps queries and passages into a shared embedding space for semantic matching, which is helpful when relevant evidence is phrased differently from the query.

\textbf{Hybrid retrieval}. To leverage both lexical and semantic signals, we fuse the sparse and dense ranked lists using Reciprocal Rank Fusion (RRF) \cite{10.1145/1571941.1572114}:
\begin{equation}
\mathrm{RRF}(d) = \sum_{i=1}^{n} \frac{1}{k + r_i(d)}
\label{eq:rrf}
\end{equation}
where $d$ is the document, $n$ is the number of ranked lists, $r_i(d)$ is the rank of document $d$ in the $i$-th ranked list, and $k$ is a constant that dampens the contribution of the lower-ranked documents.

\subsubsection{Retrieval Strategies}
\begin{figure}[t]
    \centering
    \includegraphics[width=\columnwidth]{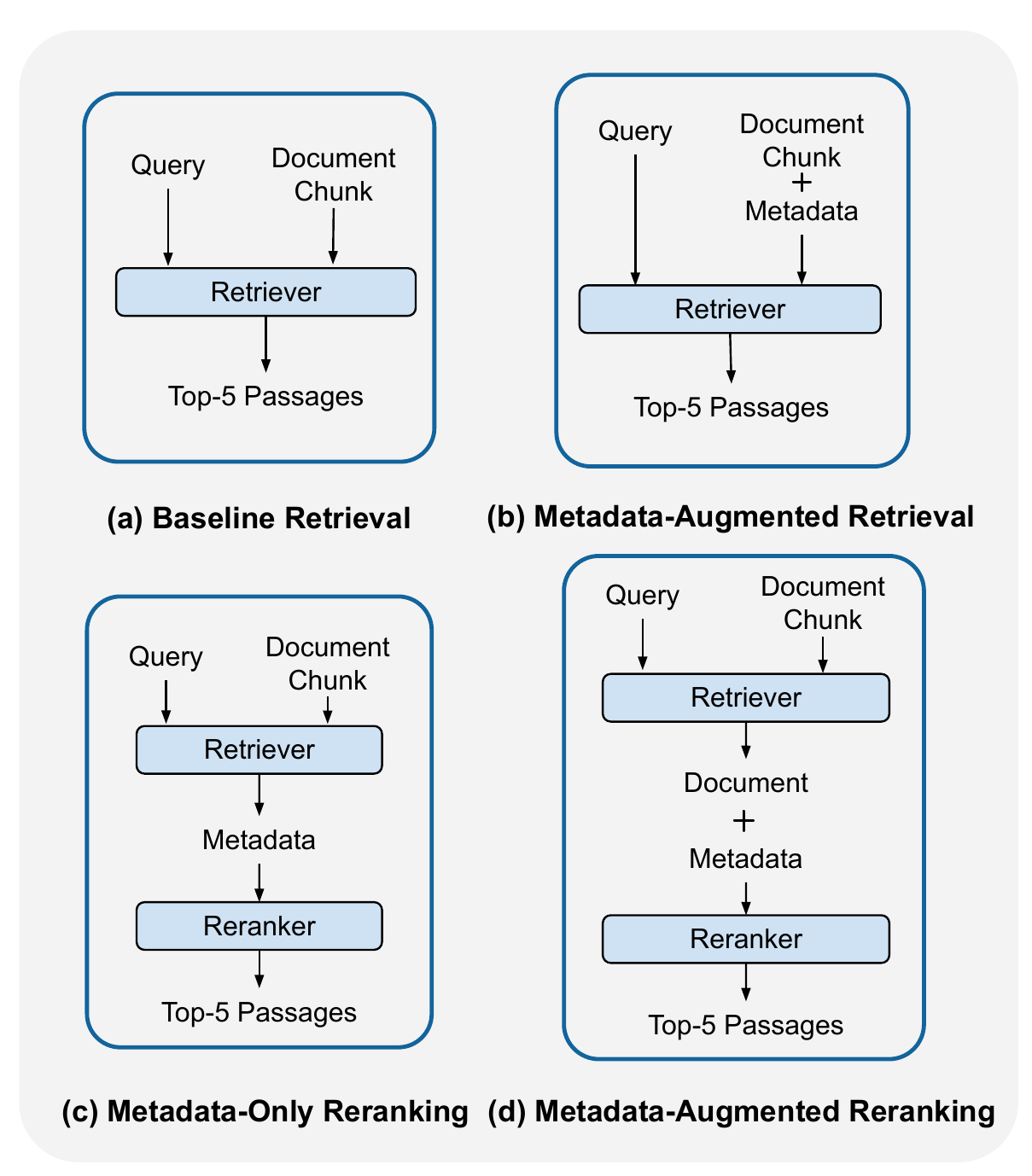}
    \caption{
    Overview of four retrieval strategies with different metadata integration stages explored in this work.  
(a) \textbf{Baseline Retrieval} retrieves top passages using only the query and document content.  
(b) \textbf{Metadata-Augmented Retrieval} integrates metadata into the document representation during retrieval.  
(c) \textbf{Metadata-Only Reranking} uses only metadata during the reranking stage after initial retrieval.  
(d) \textbf{Metadata-Augmented Reranking} incorporates both document content and metadata in the reranking stage.
}
    \label{fig:retrieval_method}
\end{figure}

Beyond first-stage retrieval, we integrate document-level metadata and a second-stage reranker to improve ranking. Metadata in TPCG is grouped into Time/Event, Person/Organization, and Content/Document fields; these fields capture signals (e.g., publication dates, presiding officials, content categories) that are often only weakly expressed in raw text but crucial for precise matching in civic or historical domains. We adopt four strategies, illustrated in Figure \ref{fig:retrieval_method}.

\textbf{Baseline Retrieval.} \
Retrieve using only the query and original document text without metadata. This provides a clean reference that relies purely on text similarity.  

\textbf{Metadata-Augmented Retrieval.} \
Append selected metadata fields to each document chunk before embedding, treating metadata as part of the content. This allows the retriever to encode, for instance, dates, roles, or categories directly into passage representations so they influence similarity at retrieval time. The retriever returns top-$k$ passages given the embeddings of query and metadata-augmented document chunks.

\textbf{Metadata-Only Reranking.} \
Incorporate metadata at the reranking stage rather than directly appended to the documents. We first retrieve the top-100 candidate passages using the original documents. Then, compute the similarity between the query and the available document-level metadata of each candidate passage. The passages are reranked based on this similarity score, and the final top-$k$ passages are returned for generation.

\textbf{Metadata-Augmented Reranking.} \
Append metadata to the original document text before computing similarity for reranking. After retrieving the candidate passages, we concatenate each document’s metadata with its original content, and then measure the similarity between this augmented text and the query to rerank the candidates. The top-$k$ passages are returned for generation.

By comparing these strategies, we aim to quantify the contribution of metadata at both embedding and reranking stages, and to better understand how different integration points influence retrieval effectiveness for historical information retrieval.

\subsection{Generation}

We use GPT-4o \cite{openai2024gpt4ocard} to produce answers conditioned on the retrieved passages. The goal is to leverage an LLM to aggregate information dispersed across multiple relevant passages into a fluent natural-language response.

At inference time, each query is paired with the top-5 retrieved passages and any available metadata, which together serve as the external knowledge context for generation. The model is instructed to ground its answer strictly in the provided materials and to avoid introducing external knowledge not mentioned in the documents. When multiple passages support the same fact, the model is encouraged to prioritize such corroborated information. If none of the provided materials is relevant to the query, the model is instructed to respond with “I don’t know”. The full generation prompt is detailed in Appendix \ref{generation_prompt}.

\subsection{Evaluation}

We evaluate both retrieval performance and end-to-end RAG quality. For retrieval evaluation, we report Recall@k, which measures the ratio of relevant passages that appear in the top-$k$ retrieved results for each query:
\begin{equation}
\text{Recall@k} = \frac{1}{N} \sum_{i=1}^{N} \mathbb{I} \left( \text{Relevant}_i \in \text{Top-}k \right)
\end{equation}
where $N$ is the number of relevant passages for the query, $\mathbb{I}(\cdot)$ is the indicator function, $\text{Relevant}_i$ is the $i^{th}$ relevant passage, and $\text{Top-}k$ denotes the top-$k$ retrieved passages. The average Recall@k across all queries yields the overall retrieval performance.


For generation quality, we employ Gemini-2.5-Pro \cite{comanici2025gemini25pushingfrontier} as an evaluator following \cite{chiang-lee-2023-large}. The evaluator is given the golden passage, the retrieved top-5 passages, and the answer from GPT-4o. The complete evaluation prompt is provided in Appendix \ref{evaluation_prompt}. It consists of three scoring dimensions: groundedness, relevance, and hallucination.

\textbf{Groundedness.} \
Assesses whether the generated answer correctly incorporates information directly supported by the golden passage. Each distinct atomic fact from the golden passage that appears correctly in the answer receives one point.

\textbf{Relevance.} \
Evaluates whether the answer includes additional information present in other retrieved passages consistent with the golden passage. Each relevant atomic fact receives one point.

\textbf{Hallucination.} \
Penalizes content that is unsupported or irrelevant. For each hallucinated statement or extraneous detail that is neither aligned with the golden passage nor substantiated by the retrieved passages, one point is deducted.

\begin{table*}[ht]
\small
\centering
\begin{tabular*}{\linewidth}{@{\extracolsep{\fill}}lcccc}
\toprule
\textbf{Question Type} & \textbf{Subcategory} & \textbf{Groundedness ↑} & \textbf{Relevance ↑} & \textbf{Hallucination ↑} \\
\midrule
All Questions & - & 2.9769 & 1.0578 & -0.6821 \\
\midrule
\multirow{2}{*}{Question Complexity} 
    & Single-hop & 2.8600 \textcolor{darkred}{(-0.1169)} & 0.8700 \textcolor{darkred}{(-0.1878)} & -0.5600 \textcolor{darkgreen}{(+0.1221)} \\
    & Multi-hop & 3.1370 \textcolor{darkgreen}{(+0.1601)} & 1.3151 \textcolor{darkgreen}{(+0.2573)} & -0.8493 \textcolor{darkred}{(-0.1672)} \\
\midrule
\multirow{6}{*}{Entity Focus} 
    & People & 3.2174 \textcolor{darkgreen}{(+0.2405)} & 1.0870 \textcolor{darkgreen}{(+0.0292)} & -0.5217 \textcolor{darkgreen}{(+0.1604)} \\
    & Event & 3.4600 \textcolor{darkgreen}{(+0.4831)} & 1.2200 \textcolor{darkgreen}{(+0.1622)} & -0.5800 \textcolor{darkgreen}{(+0.1021)} \\
    & Time & 1.3478 \textcolor{darkred}{(-1.6291)} & 0.4783 \textcolor{darkred}{(-0.5795)} & -0.9565 \textcolor{darkred}{(-0.2744)} \\
    & Place & 1.8636 \textcolor{darkred}{(-1.1133)} & 1.2273 \textcolor{darkgreen}{(+0.1695)} & -0.7727 \textcolor{darkred}{(-0.0906)} \\
    & Item & 2.5625 \textcolor{darkred}{(-0.4144)} & 0.1875 \textcolor{darkred}{(-0.8703)} & -0.5625 \textcolor{darkgreen}{(+0.1196)} \\
    & Multi-entity & 3.9744 \textcolor{darkgreen}{(+0.9975)} & 1.4359 \textcolor{darkgreen}{(+0.3781)} & -0.7436 \textcolor{darkred}{(-0.0615)} \\
\midrule All Questions (Oracle) & - & 4.4104 & 0.2312 & -0.2601 \\
\bottomrule
\end{tabular*}
\caption{
RAG evaluation by Query Type on the Fort Zeelandia dataset. The table reports average scores for three evaluation metrics: \textbf{Groundedness} (incorporates gold passage information), \textbf{Relevance} (integrates relevant passages information), and \textbf{Hallucination} (including irrelevant information). For all three metrics, higher values indicate better performance. Since Hallucination scores are negative, a value closer to zero reflects fewer hallucinations. All values are compared against the "All Questions" row. Colored deltas in parentheses indicate the difference from the average: green for improvement and red for decline. The Oracle row denotes the upper bound of the LLM’s performance when directly given the gold passages. An evaluation example can be found in Appendix \ref{fort_eval_exp}.}
\label{tab:fort_rag_score}
\end{table*}

\begin{table*}[t]
\small
\centering
\begin{tabular*}{\linewidth}{@{\extracolsep{\fill}}lcccc}
\toprule
\textbf{Integration Stage} & \textbf{Metadata Type} & \textbf{Groundedness ↑} & \textbf{Relevance ↑} & \textbf{Hallucination ↑} \\
\midrule
Baseline & - & 0.7321 & 0.8571 & -0.2500 \\
\midrule
\multirow{3}{*}{Metadata-Augmented Retrieval} 
& Time/Event & 1.0893 \textcolor{darkgreen}{(+0.3572)} & 1.0000 \textcolor{darkgreen}{(+0.1429)} & -0.2857 \textcolor{darkred}{(-0.0357)} \\
& Person/Organization & 1.1786 \textcolor{darkgreen}{(+0.4465)} & 0.7321 \textcolor{darkred}{(-0.1250)} & -0.2679 \textcolor{darkred}{(-0.0179)} \\
& Document/Content & 2.1429 \textcolor{darkgreen}{(+1.4108)} & 1.2500 \textcolor{darkgreen}{(+0.3929)} & -0.3214 \textcolor{darkred}{(-0.0714)} \\
\midrule
\multirow{3}{*}{Metadata-Only Reranking} 
& Time/Event & 0.3393 \textcolor{darkred}{(-0.3928)} & 1.0000 \textcolor{darkgreen}{(+0.1429)} & -0.4821 \textcolor{darkred}{(-0.2321)} \\
& Person/Organization & 0.5714 \textcolor{darkred}{(-0.1607)} & 0.6071 \textcolor{darkred}{(-0.2500)} & -0.2857 \textcolor{darkred}{(-0.0357)} \\
& Document/Content & 1.5893 \textcolor{darkgreen}{(+0.8572)} & 1.8571 \textcolor{darkgreen}{(+1.0000)} & -0.3393 \textcolor{darkred}{(-0.0893)} \\
\midrule
\multirow{3}{*}{Metadata-Augmented Reranking} 
& Time/Event & 1.2679 \textcolor{darkgreen}{(+0.5358)} & 1.0357 \textcolor{darkgreen}{(+0.1786)} & -0.6250 \textcolor{darkred}{(-0.3750)} \\
& Person/Organization & 0.9821 \textcolor{darkgreen}{(+0.2500)} & 1.1071 \textcolor{darkgreen}{(+0.2500)} & -0.6250 \textcolor{darkred}{(-0.3750)} \\
& Document/Content & 1.3750 \textcolor{darkgreen}{(+0.6429)} & 1.0536 \textcolor{darkgreen}{(+0.1965)} & -0.5357 \textcolor{darkred}{(-0.2857)} \\
\midrule
Oracle & - & 3.6964 & 0.0179 & -0.0714 \\
\bottomrule
\end{tabular*}
\caption{
RAG evaluation by Metadata Integration Strategies on the TPCG dataset. The table reports average scores across the three evaluation metrics.  All rows are compared to the Baseline Retrieval, values in the parentheses indicate the improvement or decline. The Oracle row denotes the upper bound of the LLM’s performance when directly given the gold passages. Two evaluation examples can be found in Appendix \ref{tpcg_eval_exp}.
}

\label{tab:rag_score}
\end{table*}

\section{Experimental Setup}
In our experiments, each document is segmented into chunks of 512 tokens with an overlap of 128 tokens to preserve contextual continuity. For direct retrieval methods, where reranking is not applied, both BM25 and BGE-M3-based approaches are configured to return the top 5 most relevant passages (i.e., top-k = 5). The hybrid method independently retrieves 5 passages using both the sparse and dense retrievers, then combines the two ranked lists using RRF, setting $k=60$, to produce the final top-5 results. For experiments involving reranking, we first retrieve the top-100 candidate passages and then apply reranking using BGE-reranker \cite{bge_embedding} to select the final top-5 results. In the reranking scenario, the hybrid approach similarly retrieves 100 passages from each retriever before merging and reranking. We do not perform any retriever and reranker tuning; all retrievers and reranker are used off-the-shelf.

For Fort Zeelandia and its query set, we use passages retrieved by a hybrid retriever with baseline retrieval. For TPCG and the associated query set, we fix the retriever to BM25 and evaluate the impact of different metadata integration stages and types on answer quality. GPT-4o is used to generate answers with the retrieved passages, and Gemini 2.5 Pro is used as an independent evaluator.

\section{Results}
Figure \ref{fig:rag_overview} illustrates the RAG pipeline and its key components at each stage. To evaluate the applicability of the RAG system on historical materials, we conduct experiments using Fort Zeelandia, TPCG, and their query sets. Our study examines how different retrieval strategies, query characteristics, and metadata integration approaches affect overall system performance. The evaluation focuses on multiple dimensions, including the ability to leverage accurate context and the extent of hallucinations.

\subsection{Overall RAG Results}
Tables \ref{tab:fort_rag_score} and \ref{tab:rag_score} show the overall RAG results on the Fort Zeelandia and TPCG datasets. In Table \ref{tab:rag_score}, Metadata-Augmented Retrieval with early Document/Content metadata achieves the highest groundedness, with a significant increase of 1.4108 over the baseline. Appendix \ref{significance_test} details the significance tests for various retrieval methods. Performance also varies by query type: event-related queries benefit most, with groundedness up 0.4831, relevance by 0.1622, and hallucinations reduced 0.1021. These findings indicate that RAG effectiveness depends on query characteristics and is strengthened by metadata-augmented retrieval, though hallucinations persist even with oracle passages, highlighting a key limitation. 

\subsection{RAG Results}
This section takes a deeper dive into two key factors that critically influence RAG performance at the Input and Retrieval stages: query type and use of document-level metadata. Specifically, we analyze how different query types affect accuracy, relevance, and hallucination. Additionally, we examine the impact of metadata integration at different stages of retrieval and reranking, considering multiple metadata types. This analysis highlights which combinations of query characteristics and metadata strategies yield the most reliable and accurate outputs for historical open-ended QA tasks.

\textbf{1) Different Query Types}
Table \ref{tab:fort_rag_score} illustrates RAG performance across query types. Multi-hop and Multi-entity questions are high-risk: when successful, groundedness increases by 0.1601 and 0.9975, and relevance by 0.2573 and 0.3781, but hallucination worsens by -0.1672 and -0.0615, highlighting a trade-off between complexity and reliability. People- and event-focused queries are more stable, achieving gains in groundedness and relevance with lower hallucination. Time-focused queries are the most challenging, with groundedness and relevance decreasing by 1.6291 and 0.5795, alongside worse hallucination, indicating that temporal reasoning remains a key bottleneck.

\textbf{2) Different Metadata Integration Strategies}
Table \ref{tab:rag_score} presents the evaluation scores across three dimensions for the open-ended question answering task, focusing on the key factor Metadata, using TPCG and its query set. 

Overall, Metadata-Augmented Retrieval proves the most reliable approach, improving groundedness and relevance with minimal worsening in hallucination. By contrast, reranking strategies show mixed results: Metadata-Only Reranking underperforms the baseline, while Metadata-Augmented Reranking achieves gains in retrieval quality but at the cost of greater hallucination, making it less stable. Across all strategies, Document/Content metadata emerges as the most effective type, underscoring its importance for enhancing the system.

\subsection{Ablation Study of Retrieval Results}
\begin{figure}[t]
    \centering
    \includegraphics[width=\columnwidth]{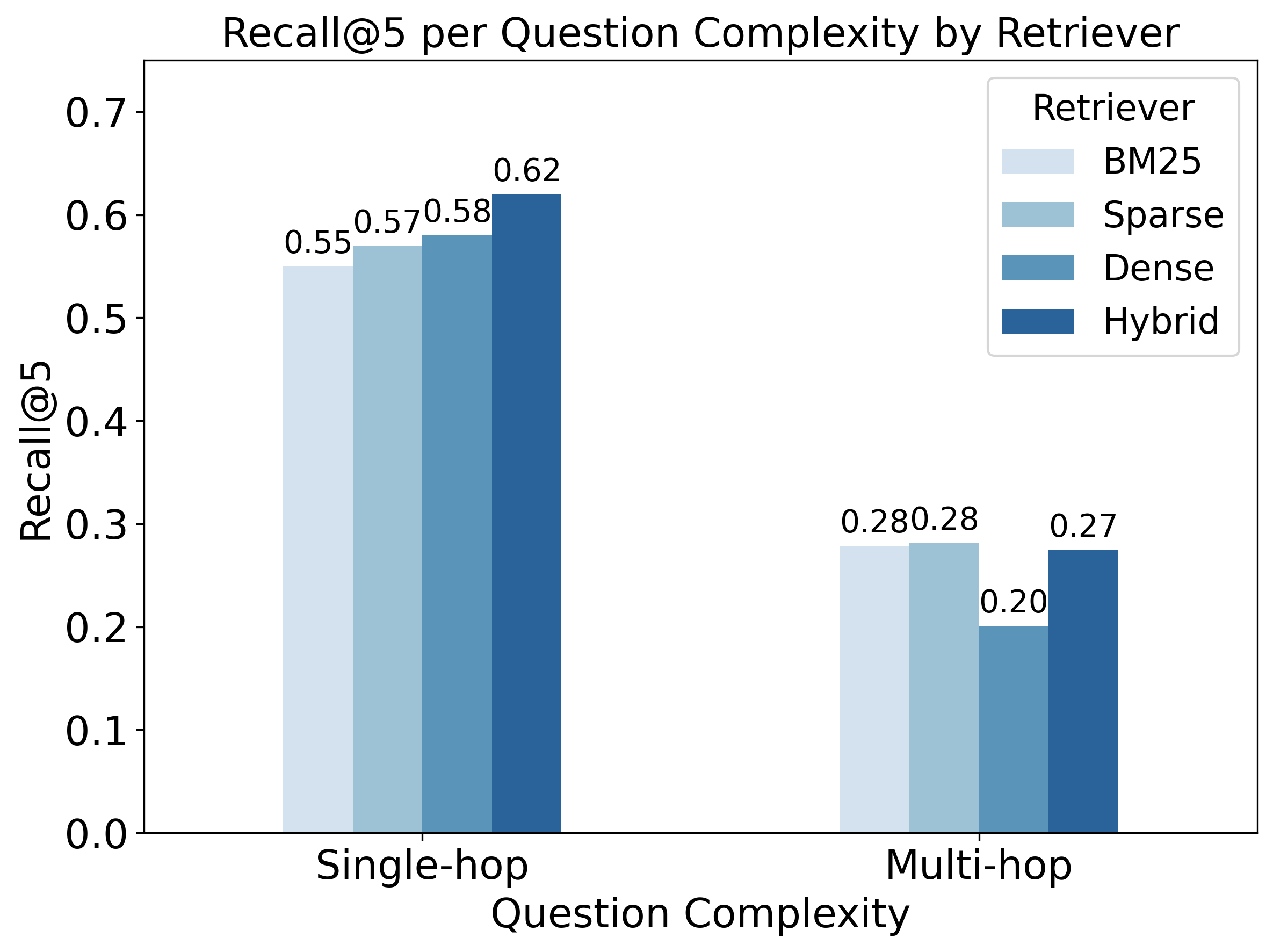}
    \caption{Fort Zeelandia Dataset Recall@5 per Question Complexity by Retriever}
    \label{fig:result_qustion_complexity}
\end{figure}

\begin{figure}[t]
\centering
\includegraphics[width=\columnwidth]{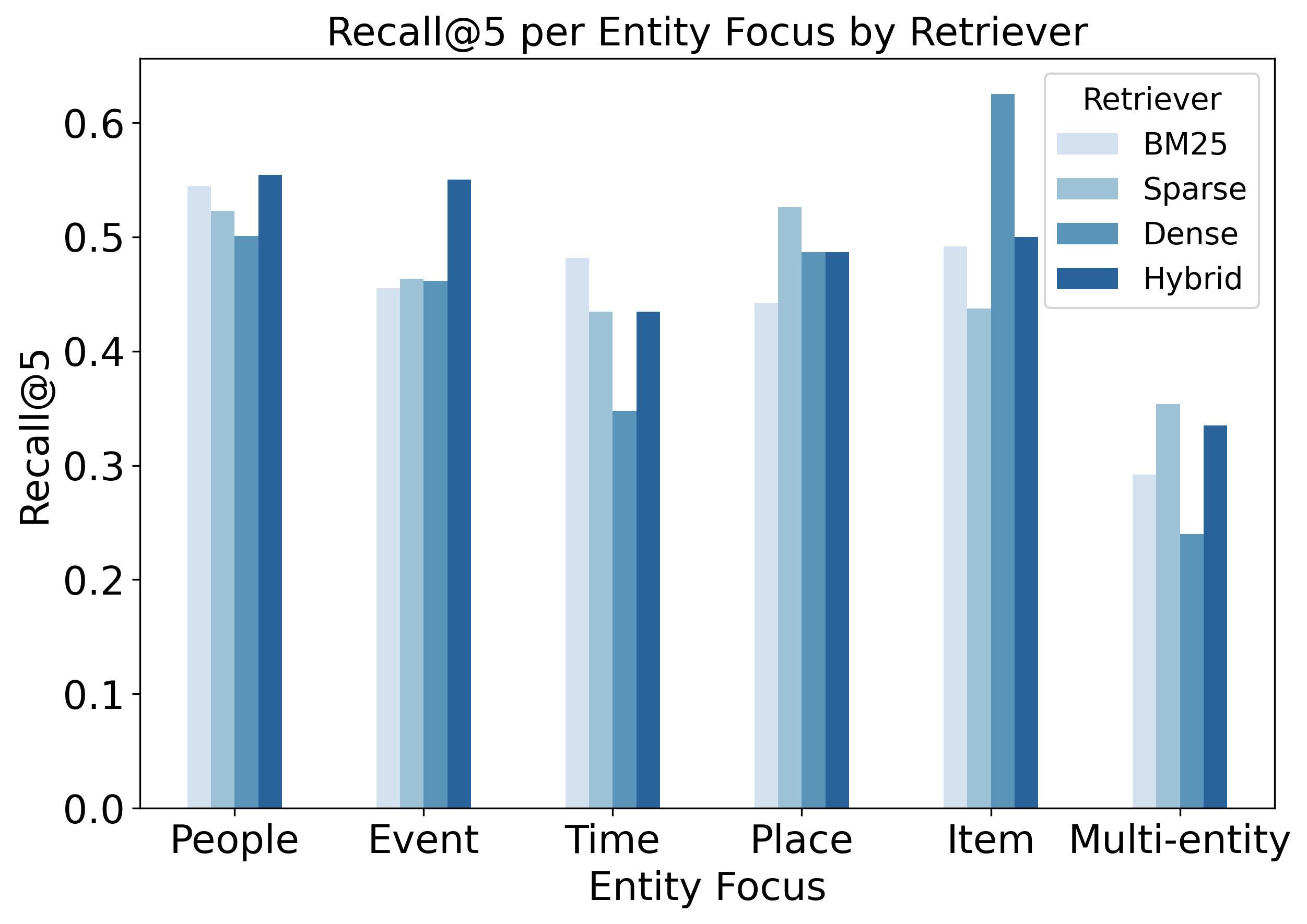}
\caption{Fort Zeelandia Dataset Recall@5 per Entity Focus by Retriever}
\label{fig:result_entity_focus}
\end{figure}

\begin{figure*}[t]
    \centering
    \includegraphics[width=\linewidth]{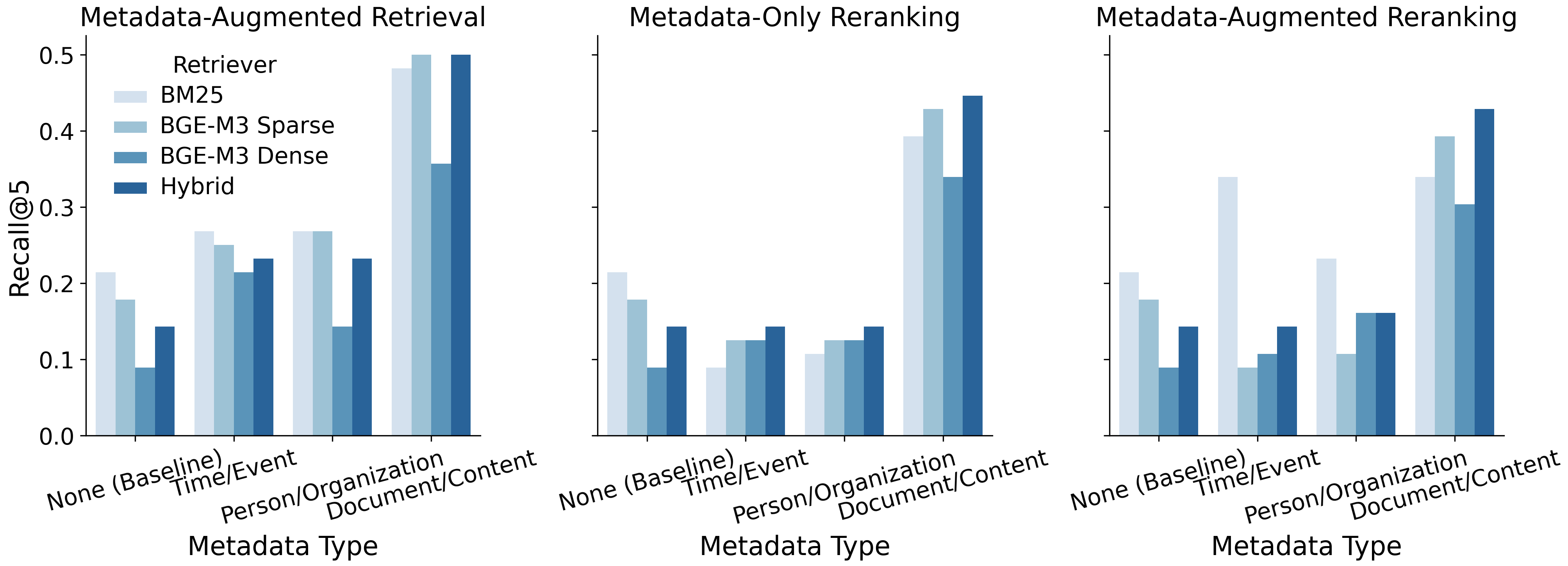}
    \caption{TPCG retrieval performance across different metadata integration stages and metadata types. Left: Metadata-Augmented Retrieval performance across different metadata types. Center: Performance of Metadata-Only Reranking across different metadata types. Right: Retrieval performance of Metadata-Augmented Reranking across different metadata types.}
    \label{fig:tpcg_retrieval_performance}
\end{figure*}

In this section, we take a closer look at the Retrieval stage of the RAG pipeline. Since RAG fundamentally relies on retrieved documents as the foundation for generating answers, understanding retrieval effectiveness is critical to interpreting overall system performance. By analyzing how different retrieval strategies, query types, and metadata integration methods influence the quality of retrieved context, we can better identify the factors that drive successes and failures in retrieval. 

\textbf{1) Retrieval with Query-level Metadata}
We investigate the impact of query types on retrieval performance using query-level metadata, focusing on query complexity and entity focus.

\textbf{Different Question Complexity.} \
To gain deeper insight into RAG performance across varying query complexity, we further analyze the retrieval results on the Fort Zeelandia dataset. Figure \ref{fig:result_qustion_complexity} presents Recall@5 scores comparing single-hop and multi-hop questions across different retrievers. For single-hop questions, Recall@5 scores are roughly twice as high as for multi-hop questions, corresponding to a lower tendency for hallucination. In contrast, retrievers achieve Recall@5 of at most only 0.28 for multi-hop queries, increasing the likelihood of hallucinated responses.

Notably, despite the lower recall, multi-hop and multi-entity questions still achieve higher groundedness and relevance, suggesting that the LLM is capable of performing multi-step reasoning when appropriate context is provided. 

\textbf{Different Entity Focus.} \
We analyze retrieval performance across different entity focuses to better understand its impact on RAG outcomes. Figure \ref{fig:result_entity_focus} presents Recall@5 scores for People, Event, Time, Place, Item, and Multi-entity questions. For the hybrid retriever used in the RAG pipeline for Fort Zeelandia, performance is notably higher for People- and Event-focused questions, with Recall@5 around 0.55, corresponding to better-controlled hallucination. In contrast, Time- and Multi-entity questions exhibit lower retrieval performance, with Recall@5 of 0.43 and 0.33, respectively, which aligns with increased hallucination.

Considering both RAG scores and retrieval results, we find that although retrieval for Time-focused questions is slightly better than for Multi-entity queries, the system achieves higher overall evaluation scores on Multi-entity questions. This indicates that the LLM can generate high-quality answers for Multi-entity queries even with partial or imperfect context. In contrast, despite adequate retrieval for Time-focused questions, generation performance remains poor, highlighting that time-sensitive reasoning constitutes a key limitation of the LLM rather than retrieval.

\textbf{2) Retrieval with Document-level Metadata}
We examine the role of document-level metadata in the retrieval process, focusing on metadata type and integration stage.

\textbf{Different Metadata Type.} \
Figure \ref{fig:tpcg_retrieval_performance} compares TPCG retrieval performance across different retrievers and metadata types: Time/Event, Person/Organization, and Document/Content, at each integration stage, arguing how metadata affects RAG performance. Document/Content metadata provides the largest improvement over the baseline across all strategies, achieving recall scores roughly twice those of the other types, with the highest around 0.5 under the Metadata-Augmented Retrieval setting. This enhanced retrieval supplies essential context to the LLM, improving answer quality and boosting groundedness and relevance, as shown in Table \ref{tab:rag_score}. In contrast, Time/Event and Person/Organization metadata exhibit variable effectiveness across integration stages and are insufficient alone for effective reranking, a trend also reflected in the RAG evaluation scores.

\textbf{Different Metadata Integration.} \
Figure \ref{fig:tpcg_retrieval_performance} also illustrates retrieval performance across different metadata integration stages. Metadata-Augmented Retrieval consistently outperforms the baseline across all retrievers and metadata types. For BM25, which is used for TPCG, recall increases from 0.21 to 0.48, indicating that integrating metadata directly into document embeddings during retrieval enables the most effective use of structured information.

In contrast, Metadata-Only Reranking produces only modest gains and sometimes underperforms the baseline; for BM25, recall drops from 0.21 to 0.08, suggesting that metadata applied solely at the reranking stage is insufficient. Metadata-Augmented Reranking yields mixed results: while recall generally improves over the baseline, gains are smaller than those of Metadata-Augmented Retrieval, leading to greater instability in generation.
\section{Conclusion}
This study investigates the application of RAG to historical open-ended question answering using two Traditional Chinese historical datasets, Fort Zeelandia and TPCG, along with query sets. 
By examining the impact of query types and metadata integration strategies on retrieval and end-to-end RAG, we show that early-stage metadata integration substantially enhances performance. Our results also reveal persistent challenges: hallucinations are frequent during generation, and temporal or multi-hop queries are particularly difficult because of the low retrieval recall. 
These findings inform future humanities-focused RAG research and underscore the need for robust retrieval strategies in historical and Traditional Chinese contexts.
\section*{Acknowledgments}
We would like to express our sincere thanks to the National Science and Technology Council (NSTC), Taiwan, for funding this research project under Grant No. NSTC 113-2740-H-002-001-MY3, “TAIHUCAIS: TAIwan HUmanities Conversational AI Knowledge Discovery System”. The support has been instrumental in enabling the study.

\bibliography{rocling2025}
\bibliographystyle{acl_natbib}
\clearpage
\appendix

\section{Appendix}

\subsection{Fort Zeelandia Dataset Example \label{fort_data}}
\begin{figure}[h]
\centering
\fbox{
\includegraphics[width=0.45\textwidth]{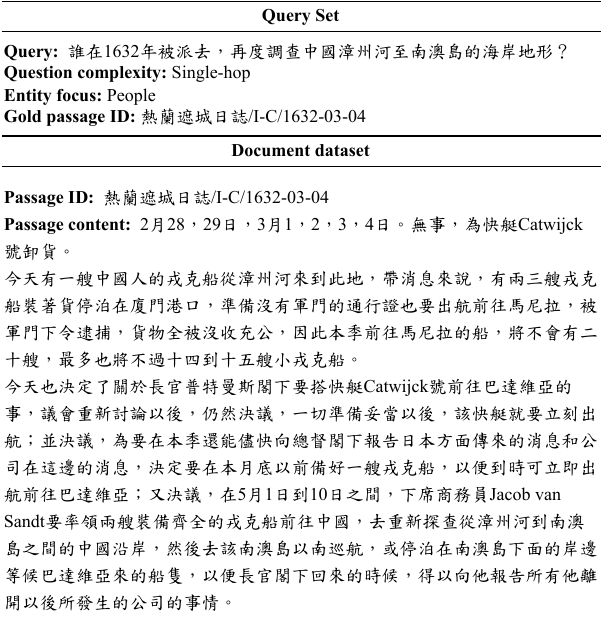}
}
\caption{A data sample of the Query Set and its relevant passage in the document dataset from the Fort Zeelandia dataset.}
\label{fig:fort_data}
\end{figure}

Figure \ref{fig:fort_data} gives an example from the Fort Zeelandia dataset.

\subsection{TPCG Dataset Example \label{tpcg_data}}
Figure \ref{fig:tpcg_data} gives an example from the TPCG dataset.

\begin{figure}[H]
\centering
\fbox{
\includegraphics[trim=0cm 18.3cm 0cm 0cm, clip, width=0.45\textwidth]{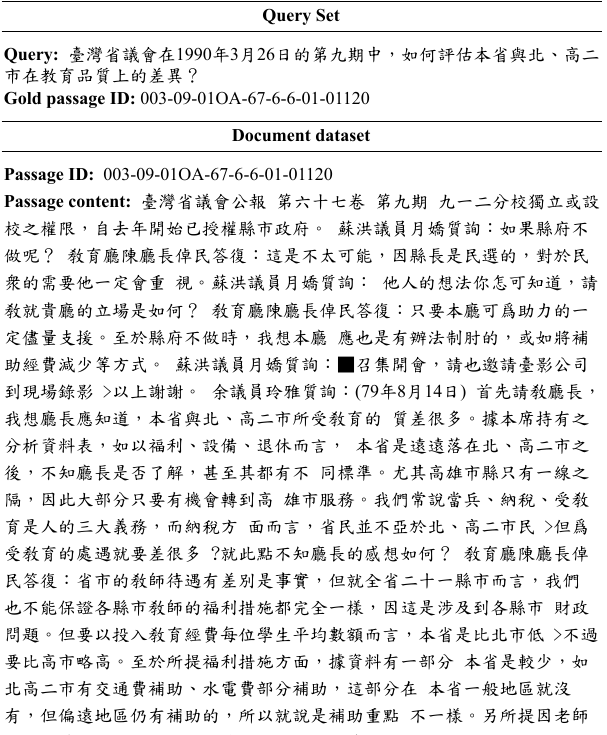}
}
\end{figure}

\begin{figure}[H]
\centering
\fbox{
\includegraphics[trim=0cm 0cm 0cm 12.3cm, clip, width=0.45\textwidth]{ROCLING2025/figure/tpcg_data.pdf}
}
\caption{A data sample of the Query Set and its relevant passage in the document dataset from the TPCG dataset. Note that some metadata fields are missing in the raw data source, such as \textbf{Decree} and \textbf{Officials}. The second half of \textbf{Passage content} is omitted for brevity.} 
\label{fig:tpcg_data}
\end{figure}

\subsection{Generation Prompt \label{generation_prompt}}
The full prompt provided to GPT-4o for response generation, given the query, retrieved passages, and available metadata, is shown in Figure \ref{fig:generation_prompt}.

\begin{figure}[H]
\centering
\fbox{
\includegraphics[trim=0cm 5.69cm 0cm 0cm, clip, width=0.45\textwidth]{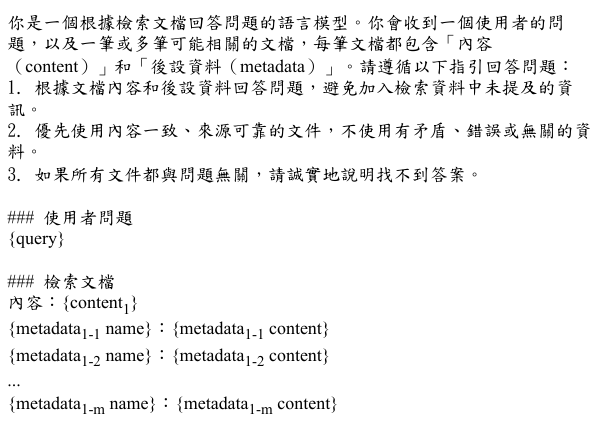}
}
\end{figure}

\begin{figure}[H]
\centering
\fbox{
\includegraphics[trim=0cm 0cm 0cm 7.1cm, clip, width=0.45\textwidth]{ROCLING2025/figure/generation_prompt.pdf}
}
\caption{RAG generation prompt to GPT-4o. Retrieved passages are numbered from $1$ to $n$, representing the $1^{st}$ retrieved passage to the $n^{th}$ retrieved passage. Metadata rows for each retrieved passage are numbered from $1$ to $m$, representing the $1^{st}$ type of metadata to the $m^{th}$ type of metadata.}
\label{fig:generation_prompt}
\end{figure}

\subsection{Evaluation Prompt \label{evaluation_prompt}}
The full prompt provided to Gemini-2.5-Pro for response evaluation, given the query, golden passages, retrieved passages, available metadata, and model response of GPT-4o, is shown in Figure \ref{fig:evaluation_prompt}.

\begin{figure}[H]
\centering
\fbox{
\includegraphics[trim=0cm 3.9cm 0cm 0cm, clip, width=0.45\textwidth]{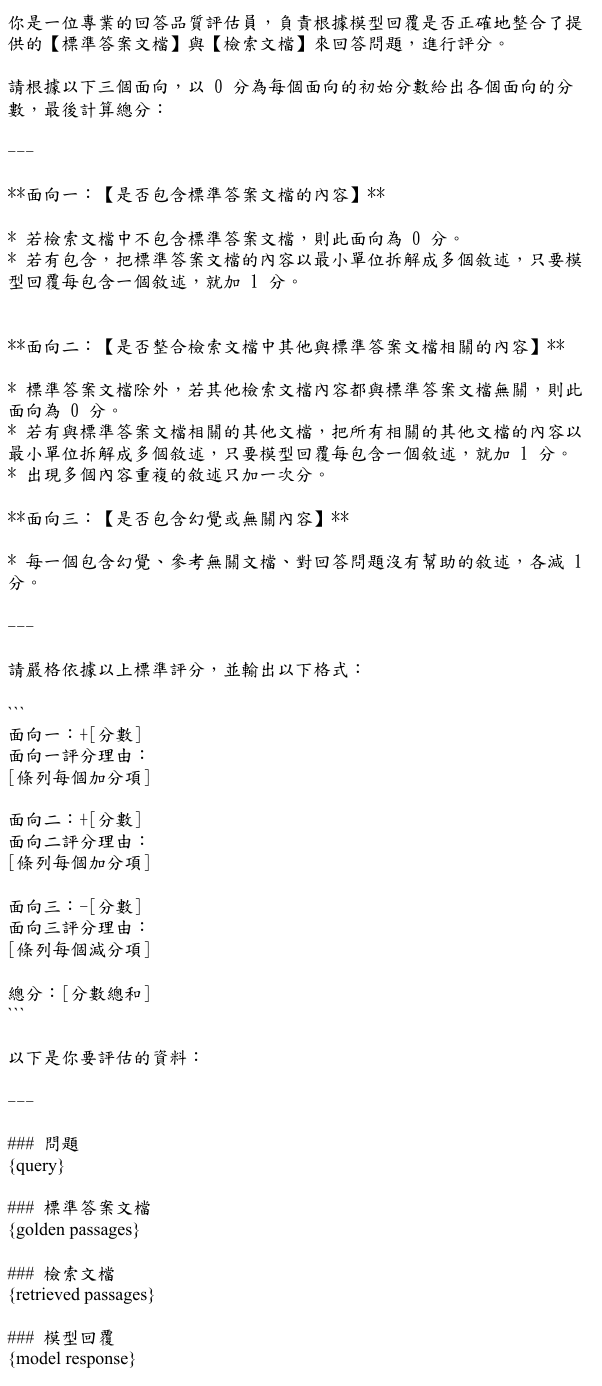}
}
\end{figure}

\begin{figure}[H]
\centering
\fbox{
\includegraphics[trim=0cm 0cm 0cm 19.6cm, clip, width=0.45\textwidth]{ROCLING2025/figure/evaluation_prompt.pdf}
}
\caption{RAG evaluation prompt to Gemini-2.5-Pro. Formats for golden passages and retrieved passages are the same as the retrieved passages in the RAG generation prompt.}
\label{fig:evaluation_prompt}
\end{figure}

\subsection{Fort Zeelandia Dataset Evaluation Example \label{fort_eval_exp}}
Figure \ref{fig:fort_eval_exp} gives a detailed example of the evaluation result on a single-hop question from the Fort Zeelandia dataset.

Focusing on the third scoring dimension of the evaluation result, we can observe that GPT-4o, which is used for model response generation, can still hallucinate, even when the golden passage is retrieved as the first retrieved passage. The hallucination may be attributed to the model's tendency not to include violence-related information from the golden passage, resulting in an incomplete response.

\subsection{TPCG Dataset Evaluation Example \label{tpcg_eval_exp}}
Figure \ref{fig:tpcg_eval_exp1} and \ref{fig:tpcg_eval_exp2} give two detailed examples of the evaluation results on the TPCG dataset.

In the first example, the model response from GPT-4o covers almost all the information in the golden passage, which is also the fifth retrieved document. However, the meeting session (in the \textbf{Title} metadata field) of the first retrieved document is wrongly linked to the golden passage and appears in the model response. This example suggests the limitation that hallucination may come from the integration of rich and complex metadata information.

In the second example, the evaluation result of the second scoring dimension shows that GPT-4o can still summarize related information from other retrieved passages even though the golden passage is not given for generation. Nonetheless, the model regards the requirements and questions, which are raised by council members, as implemented specific measures, introducing another type of hallucination due to the imprecise interpretation of retrieved passages.

\begin{figure*}
\centering
\fbox{
\includegraphics[trim=0cm 45.7cm 0cm 0cm, clip, width=0.98\textwidth]{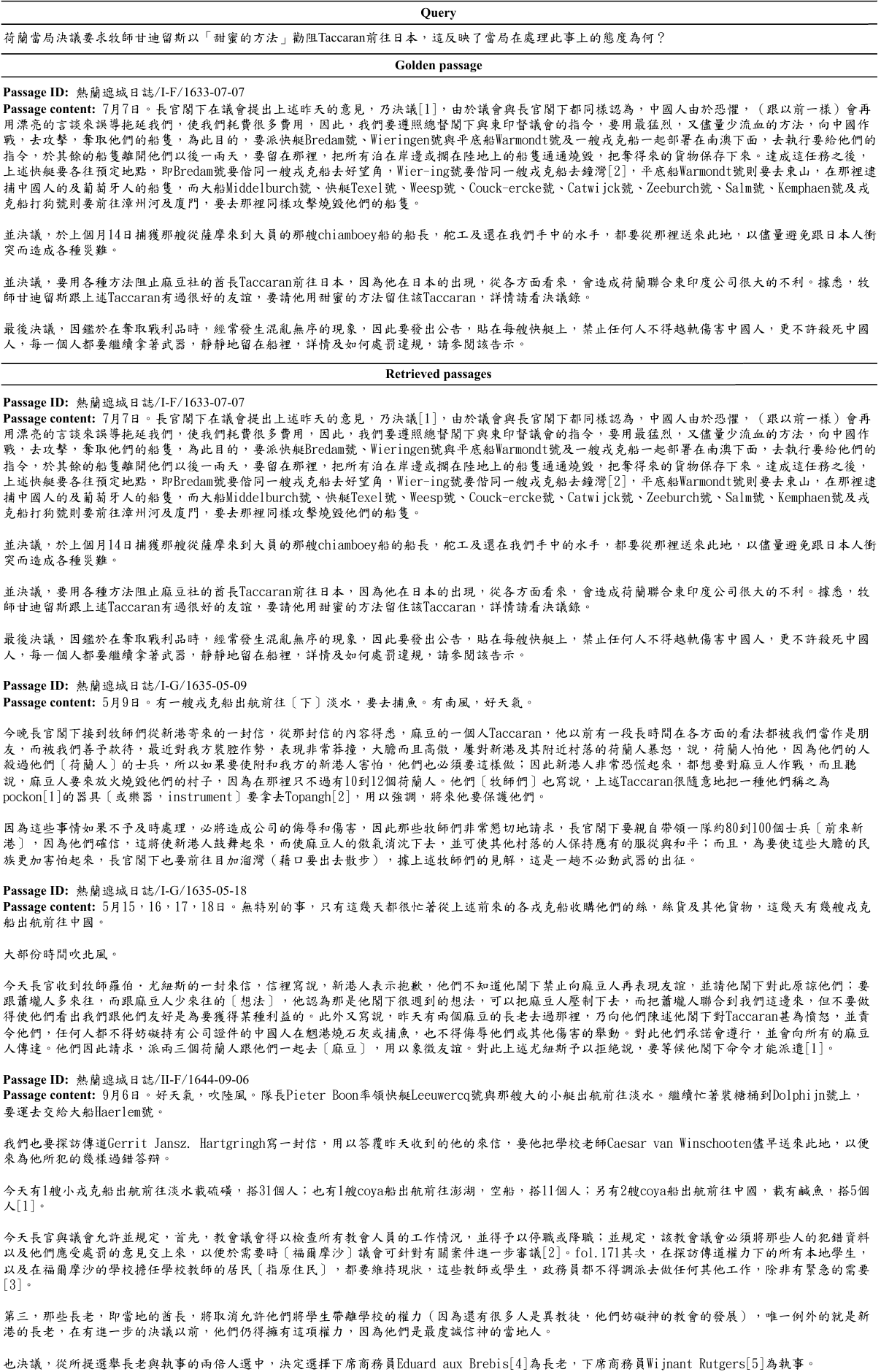}
}
\end{figure*}

\begin{figure*}
\centering
\fbox{
\includegraphics[trim=0cm 13.9cm 0cm 32cm, clip, width=0.98\textwidth]{ROCLING2025/figure/fort_eval_exp.pdf}
}
\end{figure*}

\begin{figure*}
\centering
\fbox{
\includegraphics[trim=0cm 0cm 0cm 63.5cm, clip, width=0.98\textwidth]{ROCLING2025/figure/fort_eval_exp.pdf}
}
\caption{Evaluation result on the Fort Zeelandia dataset.}
\label{fig:fort_eval_exp}
\end{figure*}

\begin{figure*}
\centering
\fbox{
\includegraphics[trim=0cm 44.9cm 0cm 0cm, clip, width=0.98\textwidth]{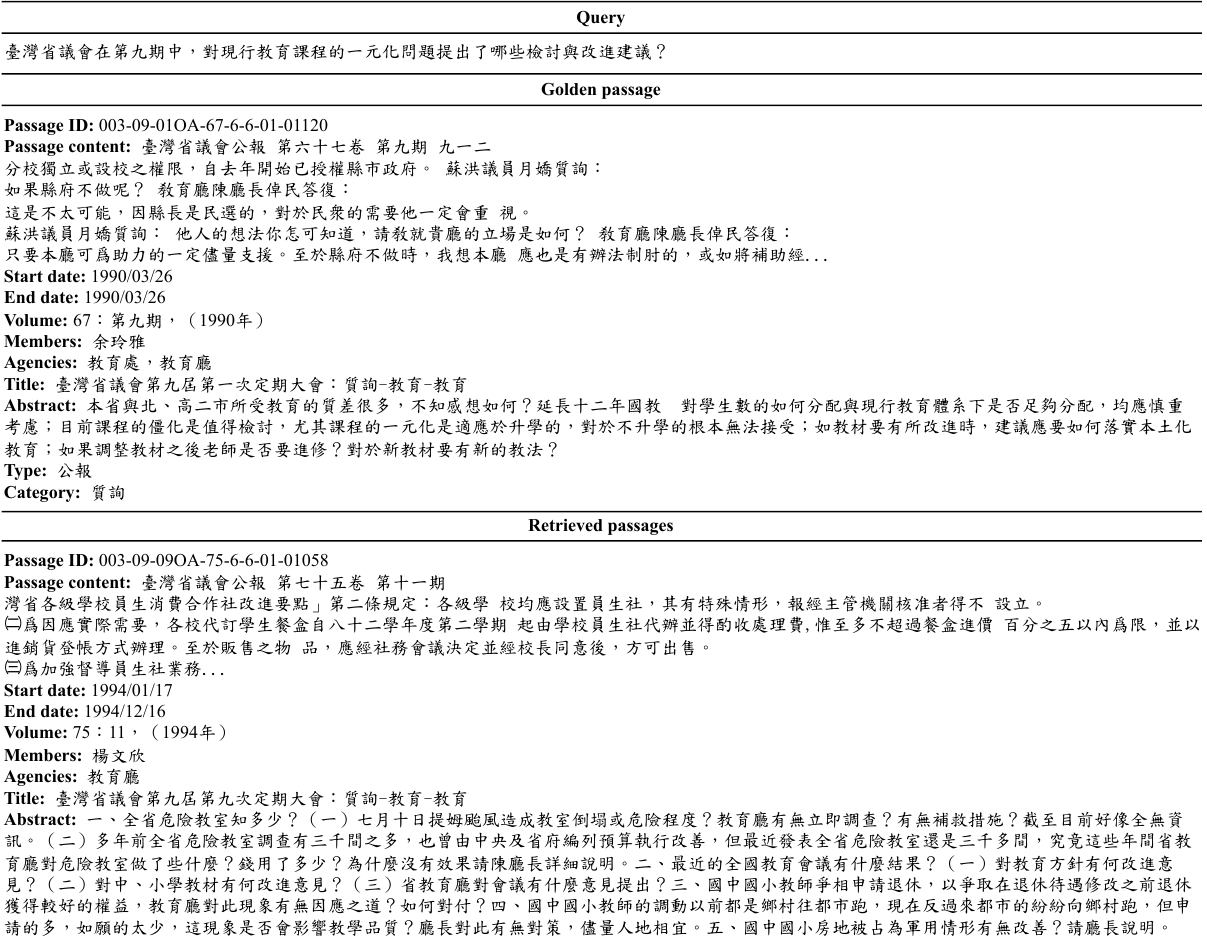}
}
\end{figure*}

\begin{figure*}
\centering
\fbox{
\includegraphics[trim=0cm 13.1cm 0cm 15.9cm, clip, width=0.98\textwidth]{ROCLING2025/figure/tpcg_eval_exp1.pdf}
}
\end{figure*}

\begin{figure*}
\centering
\fbox{
\includegraphics[trim=0cm 0cm 0cm 47.7cm, clip, width=0.98\textwidth]{ROCLING2025/figure/tpcg_eval_exp1.pdf}
}
\caption{First example of evaluation result on the TPCG dataset. For brevity, part of \textbf{Passage content} and empty metadata fields for each passage are omitted.}
\label{fig:tpcg_eval_exp1}
\end{figure*}

\begin{figure*}
\centering
\fbox{
\includegraphics[trim=0cm 38.7cm 0cm 0cm, clip, width=0.98\textwidth]{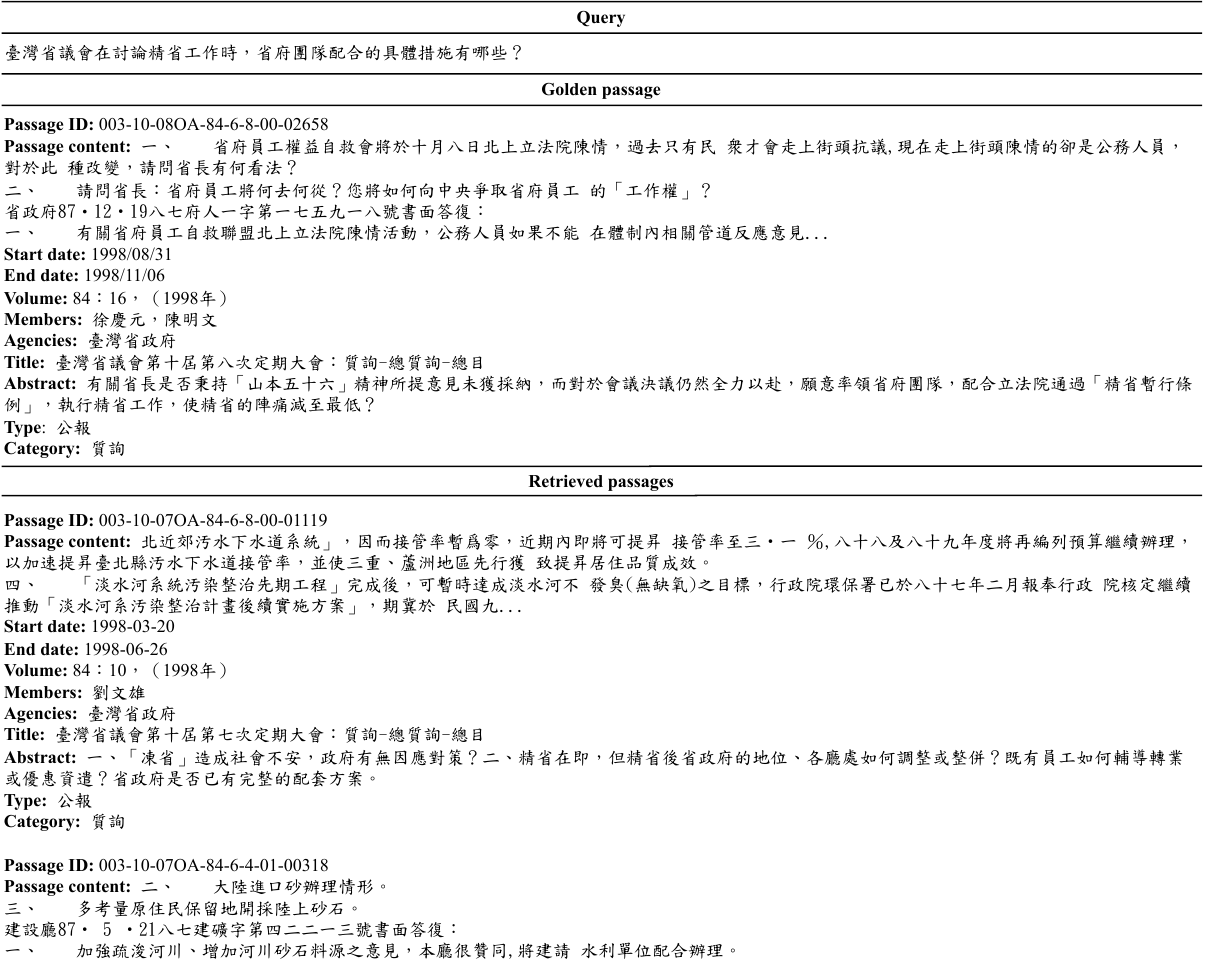}
}
\end{figure*}

\begin{figure*}
\centering
\fbox{
\includegraphics[trim=0cm 6.9cm 0cm 16.3cm, clip, width=0.98\textwidth]{ROCLING2025/figure/tpcg_eval_exp2.pdf}
}
\end{figure*}

\hspace{-0.37cm}\begin{minipage}{\textwidth}
\begin{figure}[H]
\centering
\fbox{
\includegraphics[trim=0cm 0cm 0cm 48cm, clip, width=0.98\textwidth]{ROCLING2025/figure/tpcg_eval_exp2.pdf}
}
\caption{Second example of evaluation result on the TPCG dataset. For brevity, part of \textbf{Passage content} and empty metadata fields for each passage are omitted.}
\label{fig:tpcg_eval_exp2}
\end{figure}
\end{minipage}

\hspace{-0.37cm}\begin{minipage}{\textwidth}
\begin{table}[H]
\small
\centering
\begin{tabular*}{\textwidth}{@{\extracolsep{\fill}}lcccc}
\toprule
Method & Metadata Type & Mean $\Delta$ & p-value & Significant \\
\midrule
\multirow{3}{*}{Metadata-Augmented Retrieval} 
  & Time/Event & 0.3571 & 0.0327 & \cmark \\
  & Person/Organization & 0.4464 & 0.0619 & \xmark \\
  & Document/Content & 1.1407 & 0.0007 & \cmark \\
\midrule
\multirow{3}{*}{Metadata-Only Reranking} 
  & Time/Event & -0.3929 & 0.9175 & \xmark \\
  & Person/Organization & -0.1607 & 0.7156 & \xmark \\
  & Document/Content & 0.8571 & 0.0005 & \cmark \\
\midrule
\multirow{3}{*}{Metadata-Augmented Reranking} 
  & Time/Event & 0.5357 & 0.0036 & \cmark \\
  & Person/Organization & 0.2500 & 0.1095 & \xmark \\
  & Document/Content & 0.6429 & 0.0047 & \cmark \\
\bottomrule
\end{tabular*}
\caption{Wilcoxon signed-rank test results comparing each retrieval method and metadata type against the baseline for Groundedness on TPCG. The table shows the mean difference ($\Delta$), p-value, and whether the improvement is statistically significant at $p < 0.05$.}
\label{tab:groundedness_sig}
\end{table}   
\end{minipage}

\subsection{RAG Groundedness Significance Test  \label{significance_test}}
Table \ref{tab:groundedness_sig} presents the detailed results of significance testing for the Groundedness metric. For each combination of method and metadata type, we report the mean difference compared to the baseline, the corresponding p-value from the Wilcoxon signed-rank test, and a visual indicator of statistical significance. The results show that the Document/Content metadata type provides the most substantial benefit across retrieval stages, and among the methods, Metadata-Augmented Retrieval with Document/Content metadata achieves the largest mean difference, indicating the strongest improvement over the baseline.

\end{document}